# Hybrid Optimized Back propagation Learning Algorithm For Multi-layer Perceptron

Arka Ghosh
Purabi Das School of Information Technology,
Bengal Engineering & Science University , Shibpur,
Howrah, West Bengal, India.
E-mail:-arka.besu@gmx.com

Mriganka Chakraborty
Asst. Professor
Department Of Computer Science & Engineering,
Seacom Engineering College,
Howrah, West Bengal, India.
E-mail:-cmri.net@gmail.com

## ABSTRACT
Standard neural network based on general back propagation learning using delta method or gradient descent method has some great faults like poor optimization of error-weight objective function, low learning rate, instability .This paper introduces a hybrid supervised back propagation learning algorithm which uses trust-region method of unconstrained optimization of the error objective function by using quasi-newton method .This optimization leads to more accurate weight update system for minimizing the learning error during learning phase of multi-layer perceptron.[13][14][15] In this paper augmented line search is used for finding points which satisfies Wolfe condition. In this paper, This hybrid back propagation algorithm has strong global convergence properties & is robust & efficient in practice.

## 1. INTRODUCTION
Brain is the central processing unit of .all the living intelligent beings. Brain is made by simple processing elements called neurons. These neurons reside in brain as a massively interconnected network. Inspired by this biological neural system artificial neural network model is proposed. Over past fifteen years, a view has emerged that computing based on models inspired by our understanding of the structure and function of the biological neural networks may hold the key to the success of solving intelligent tasks by machines [1] A neural network is a massively parallel distributed processor that has a natural propensity for storing experiential knowledge and making it available for use. It resembles the brain in two respects: Knowledge is acquired by the network through a learning process and interneuron connection strengths known as synaptic weights are used to store the knowledge [2].

Several models of artificial neuron has been proposed, some major models are like Mc-Culloch-Pitts model [3], Perceptron model [4], Adaline model etc. [5]. There are a handful of learning laws are also available some most important learning laws used for neural network learning are Perceptron learning law [6], Hebb's learning law [1], Delta learning law [2], Widrow and Hoff LMS learning law [1], Correlation learning law [1], Instar learning law[1] etc.

Back propagation learning method for multi-layer perceptron network is extensively used in last few decades in many fields of science and technology. Main task of this back propagation learning algorithm can be divided into two sub-objectives (a) feed forward computation and (b) back propagation of error to minimize the total learning error of the network. In this method error is controlled by tuning those synaptic weight values defined between different neural nodes. There are different major practices to optimize those weight vectors to minimize the total error of the network system, most common one among those practices is use gradient descent method in back propagation learning to optimize weight vector and minimize error accordingly. There are some major disadvantages of this gradient descent approach, one of them is stuck into local minima ,which can be mostly avoided by using a learning rate but that sometime may cause serious problem of overshooting, there also another problem of very slow convergence of the learning algorithm which severely depends upon choosing right value for learning rate[17]. For these reasons there are some more methods available to use in aid of standard back propagation learning, one of them is using Quasi-Newton optimization approach for that weight vector with respect to weight vector[18][19]. In this paper a hybrid back propagation learning method which uses quasi-newton optimization for optimized weight updating is discussed.

BFGS Method is a classical quasi-Newton method, and also is one of the most effective algorithms of the unconstrained optimization problems at present [7]. The BFGS algorithm (Nocedal and Wright, 1999) was developed independently by Broyden, Fletcher, Goldfarb, and Shanno. The basic principle in quasi-Newton methods is that the direction of search is based on an $n \times n$ direction matrix **S** which serves the same purpose as the inverse Hessian in the Newton method. This matrix is generated from available data and is contrived to be an approximation of $H^{-1}$ Furthermore, as the number of iterations is increased, **S** becomes progressively a more accurate representation of $H^{-1}$, and for convex quadratic objective functions it becomes identical to $H^{-1}$ in $n + 1$ iterations [20]. By using quasi-newton optimization(BFGS hessian update) in minimization of error of back propagation learning of multilayer perceptron networks it is proved to be very much helpful which is demonstrated in this paper.

## 2. METHODOLOGY

### 1. Back-Propagation Learning

The popularity of on-line learning for the supervised training of multilayer perceptron has been further enhanced by the development of the back-propagation For illustration of back-propagation learning please consider Figure-1.

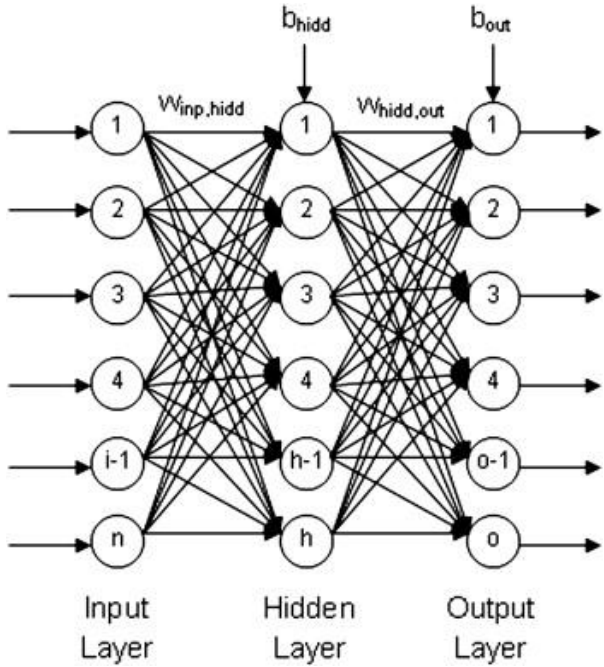

**Figure-1. Multi-layer perceptron network**

There are three major layers in a multilayer perceptron network, first layer in input layer where input signal is fed to the nodes of network (1,2,…,n), second layer is hidden layer(1,2,…,h), and third one is output layer (1,2,…,o). The entire computation of the system is done in hidden layer, from output layer provides output of the said network. There are two more things bhidd and bout, these are bias node connected with hidden layer and output layer respectively. Winp,hidd is the synaptic weight values assigned between input layer and hidden layer of the network, and Whidd,out is the synaptic weight values assigned between hidden layer and the output layer of the network. Controlling and minimization of total error of the network is done by tuning these weight values. In this paper error is as mean square error [9] between original output $T_k$ .and network simulated output $O_k$.

**Back-Propagation Algorithm (Gradient Descent)[10]**

Initialize each weight $wi$ to some small random value.

• Until termination condition is met do –>

-For each training example do –>

• Input it & compute network output $O_k$

• For each output unit k

$\delta k <- Ok(1-Ok)(Tk-Ok)$

• For each hidden unit h

$\delta h <- Oh(1-Oh)\sum_{k \in outputs} wh,k\, \delta k$

For each network weight $wi$ do –>

$wi, <- wi,+\Delta wi,j$

Where $\Delta wi, = \eta \delta j xi$,

Here the transfer function is sigmoid transfer function, it is used for its continuous nature. $\eta$ is the learning rate & $\delta$ is the gradient.

### 2. Quasi –Newton Method

The main objective of this method is to find the second order approximation of the minimum of a function. f(x). The Taylor series of f(x) is given by

$$f(x_k + \Delta x) \approx f(x_k) + \nabla f(x_k)^T \Delta x + \frac{1}{2}\Delta x^T H \Delta x$$

Where $\nabla(f(x))$ is the gradient of the objective function f(x) and H is the hessian matrix. The gradient of this approximation with respect to $\Delta x$ is defined by

$$\nabla f(x_k + \Delta x) \approx \nabla f(x_k) + H\Delta x$$

Quasi-Newton method is based on Newton method which finds the stationary point of a function where gradient of that objective function is zero. Now setting this gradient to zero,

$$\Delta x = -H^{-1}\nabla f(x_k)$$

In practice computing the hessian that many time for a given objective function is very much expensive in the terms of memory and time so BFGS update formula is used to approximate hessian H by B matrix. This hessian approximation B has to satisfy the following equation,

$$\nabla f(x_k + \Delta x) = \nabla f(x_k) + B\Delta x$$

The initial value of B is approximated by $B_0 = I$, the updated value of $x_k$ is calculated by applying Newton's step calculation using $B_k$ which is current approximation of the hessian matrix[11]

$$\Delta x_k = -\alpha B_k^{-1} \nabla f(x_k)$$

Where $\alpha$ is chosen to satisfy the Wolfe condition [12].

$$x_{k+1} = x_k + \Delta x_k$$

$\nabla f(x_k + 1)$ is the gradient at new point and

$$y_k = \nabla f(x_k + 1) - \nabla f(x_k)$$

Is necessary for computing updated hessian approximation $B_{k+1}$.

BFGS update formula is used in this paper for updating the approximate hessian matrix. The equation is given by,

$$B_{k+1} = B_k + \frac{y_k y_k^T}{y_k^T \Delta x_k} - \frac{B_k \Delta x_k (B_k \Delta x_k)^T}{\Delta x_k^T B_k \Delta x_k}$$

$$H_{k+1} = B_{k+1}^{-1} = \left(I - \frac{y_k x_k^T}{y_k^T \Delta x_k}\right)^T H_k \left(I - \frac{y_k x_k^T}{y_k^T \Delta x_k}\right) + \frac{\Delta x_k \Delta x_k^T}{y_k^T \Delta x_k}$$

In this paper this BFGS update equation is used for successive weight update, here objective function is defined by the first order partial derivative of the error cost function with respect to the corresponding weight. So, the proposed back-propagation algorithm is,

**Back-Propagation Algorithm (Quasi-Newton with BFGS Update)**

Initialize each weight $wi$ to some small random value.

• Until termination condition is met do ->

-For each training example do ->

• Input it & compute network output $Ok$

• For each output unit k

$\delta k <- Ok \ (1- \ )(Tk - Ok)$

• For each hidden unit h

$\delta h <- Oh \ (1 - Oh) \sum_{k \in outputs} wh, k \ \delta k$

• For each network weight $wi$ do –>

$wi, <- wi, +\Delta wi, j$

Where $\Delta w = -H-1 \nabla w \delta w$

## 3. EXPERIMENTAL RESULT

Proposed back-propagation learning algorithm is used for training purpose of a multi-layer perceptron network which is used for universal function approximation. Its performance is measured in terms of mean square error of training and testing phase. This network is used to approximate two functions,

- **Beale Function:-**

This two dimensional function computes

$$f(n,x) = (1.5 - x_0 + x_0 x_1)^2 + (2.25 - x_0 + x_0 x_1)^2 + (2.625 - x_0 + x_0 x_1)^2$$

With domain $-4.5 \leq x_i \leq 4.5$.

Here No of variables are two. In testing randomly generated values are used for these two variables then using said function the corresponding original function value is calculated then a multi-layer perceptron network learnt with proposed algorithm is used to approximate this function. Results are shown below,

The test error is: 1.3954% (Simulated data are from test data).

The training error is: 0.10709% (Simulated data are from training data).

Calculated CPU Time: 69.3736 seconds.

Figure-2 and Figure-3 depicts original beale function data and MLP simulated beale function data.

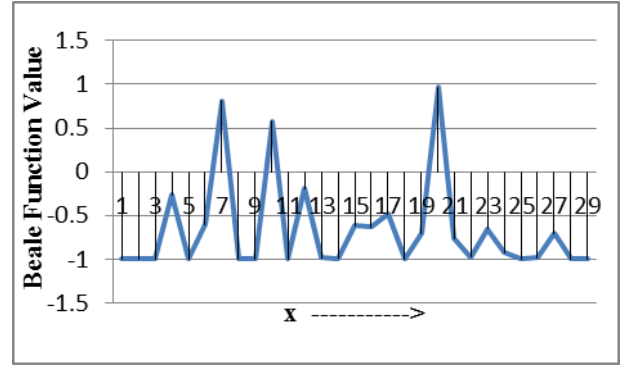

**Figure-2. Plot of original Beale function data.**

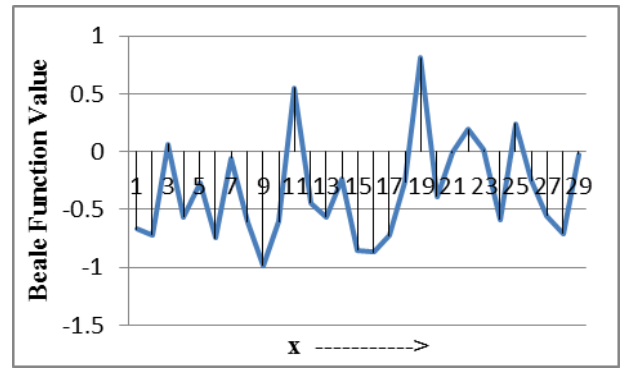

**Figure-3. Plot of MLP simulated Beale function data.**

From these two plots one can say that the original function data and the simulated one are very much close enough. Figure-5 and Figure-6 gives us clear view of regression and performance of the MLP network.

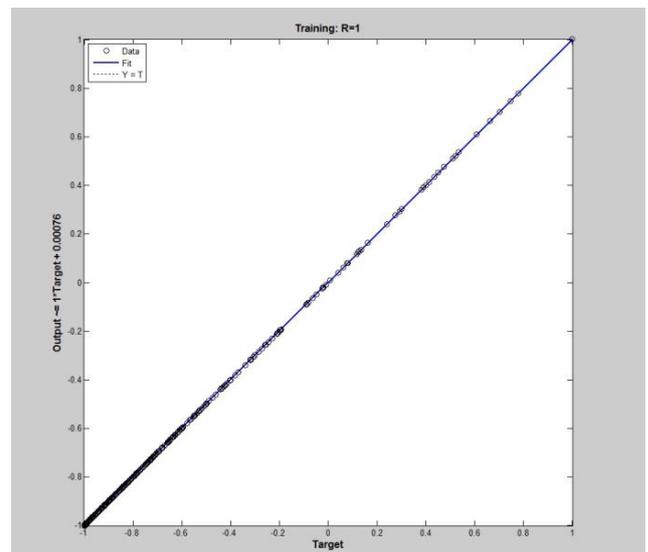

**Figure-4. Regression Plot of MLP.**

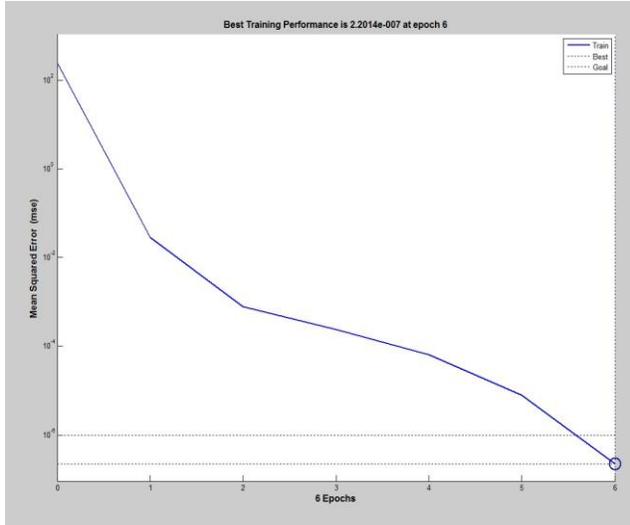

**Figure-5 .Performance(MSE) Plot of MLP.**

- **Booth Function:-**

The two-dimensional function computes

$$f(n,x) = (x_0 + 2x_1 - 7)^2 + (2x_0 + x_1 - 5)^2$$

With domain $-10 \leq x_i \leq 10$.

The test error is: 1.44% (Simulated data are from test data)

The training error is: 0.009874% (Simulated data are from training data)

Calculated CPU-Time: 70.2473seconds

Figure-6 and Figure-7 depicts original booth function data and MLP simulated booth function data.

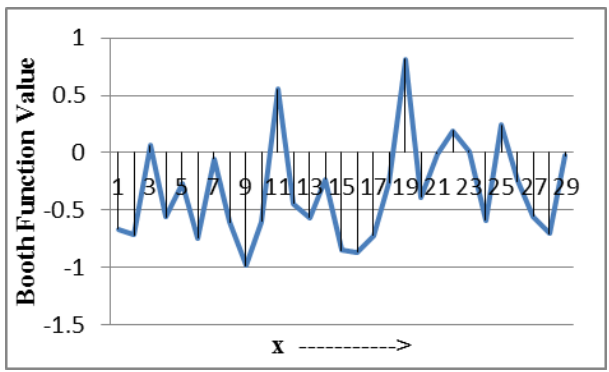

**Figure-6. Plot of Booth function data.**

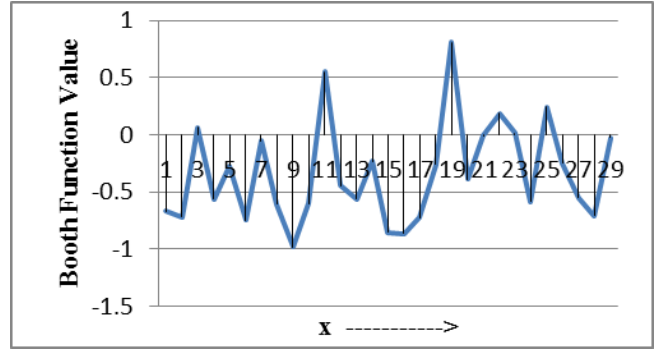

**Figure-7.Plot of MLP simulated Booth function data.**

From these two plots one can say that the original function data and the simulated one are very much close enough.Figure-8 and Figure-9 gives us clear view of regression and performance of the MLP network.

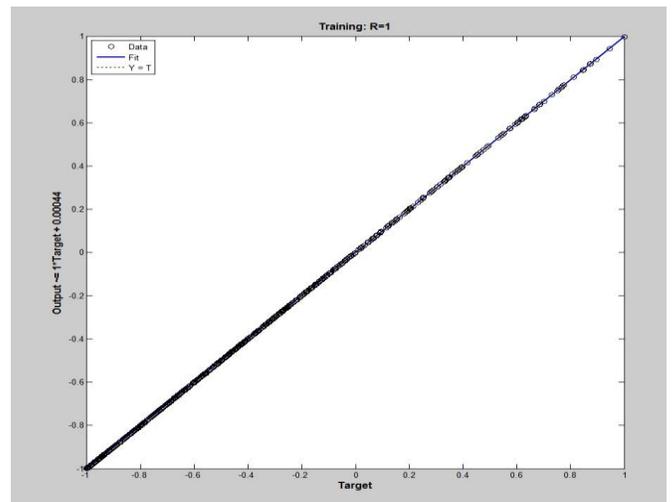

**Figure-8.Regression Plot of MLP.**

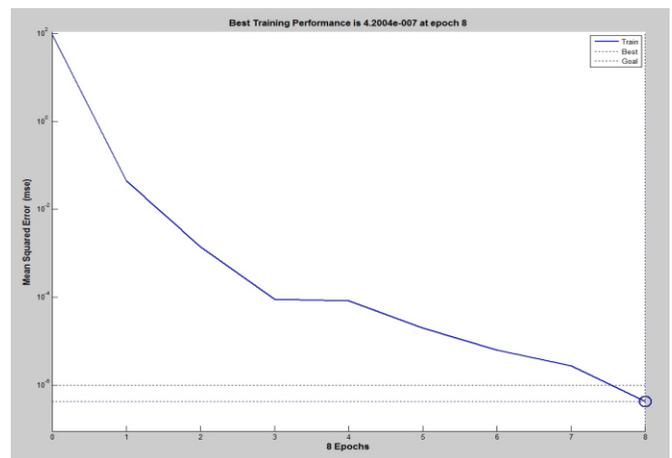

**Figure-9 .Performance (MSE) Plot of MLP**

|  | % Of Error | |
|---|---|---|
| **Training Algorithm** | **Booth Function** | **Beale Function** |
| **Proposed Algorithm** | 1.44% | 1.3954% |
| **Gradient Descent** | 13.59% | 16.77% |

**Table-1 .Comparative Performance (MSE) Analysis of Proposed Method with Existing One**

## 4. CONCLUSION

In this paper, a hybrid optimized back propagation learning algorithm is proposed for successful learning of multi-layer perceptron network .This learning algorithm, utilizing an artificial neural network with the quasi-Newton algorithm is proposed for design optimization of function approximation. The method can determine optimal weights and biases in the network more rapidly than the basic back propagation algorithm or other optimization algorithms.

Two modifications to the classical approach of the Quasi-Newton method have been presented. It was shown that the hypotheses supporting those methods are relevant and desirable in terms of convergence properties. It represents a clear gain in terms of computational time without a major increase in memory space required, making the approach suitable for large scale problems. There is also no need to adjust parameters, as in the back-propagation algorithm, which makes our algorithm very easy to use.

## 6. AUTHORS PROFILE

**Arka Ghosh** was born on 7[th] November 1990 in Burdwan , West Bengal, India. He has has received the B.Tech degree in computer science & engineering from West Bengal University of Technology, Kolkata, West Bengal, India in 2012. He is currently pursuing his post graduate studies in Information technology from Bengal Engineering & Science University, Shibpur, Howrah, West Bengal, India. His research interests includes Artificial Neural Network, Support Vector Machines, Fuzzy logic.

**Mriganka Chakraborty** was born on 11th October 1984 in Kolkata, West Bengal, India. He has received the B.Tech degree in computer science & engineering and M.Tech degree in vlsi & microelectronics from West Bengal University of Technology, Kolkata, West Bengal, India in 2006 and 2009 respectively. He is currently an Asst. Professor with the Department of Computer Science & Engineering in Seacom Engineering College, Howrah, West Bengal, India. His research interests includes Artificial Neural Network, Support Vector Machines, Particle Swarm Optimization, VLSI circuit design, Network On Chip, Physical VLSI Design techniques.